# A large-scale pipeline for LLM-assisted corpus annotation

## Variation and change in the English *consider* construction

Cameron Morin and Matti Marttinen Larsson
Paris Cité University | Stockholm University

As natural language corpora expand at an unprecedented rate, manual annotation remains a significant methodological bottleneck in corpus linguistic work. We address this challenge by presenting a scalable pipeline for automating grammatical annotation using large language models (LLMs). We demonstrate the pipeline's accessibility and effectiveness through a diachronic case study of variation in the English evaluative *consider* construction (*consider* X *as*/*to be*/Ø Y). We annotate 143,933 'consider' concordance lines from the Corpus of Historical American English (COHA) via the OpenAI API in under 60 hours, achieving 98%+ accuracy on two sophisticated annotation procedures. Subsequent analysis reveals previously undocumented genre-specific trajectories of change, enabling us to advance new hypotheses about the relationship between register formality and competing pressures of morphosyntactic reduction and enhancement. Our results suggest that LLMs can perform a range of data preparation tasks at scale with minimal human intervention, unlocking substantive research questions previously beyond practical reach.

**Keywords:** large language models, automated annotation, diachronic variation, grammatical constructions, prompt engineering

## 1. Introduction

The exponential growth of digital text corpora has fundamentally reshaped the empirical landscape of linguistics. Where researchers once worked with carefully curated datasets of thousands or tens of thousands of words, they now routinely access multi-billion-word collections spanning diverse registers, time periods, and language varieties. Web-based corpora such as the News on the Web (NOW) corpus (Davies, 2016), the TenTen family (Kilgarriff et al., 2014), and social media datasets (Grieve et al., 2018) provide unprecedented opportunities to study linguistic variation at scale, to detect emerging changes in real time, and to investigate phenomena that occur with such low frequency that they would be invisible in smaller datasets. This transformation has been particularly consequential for variationist and diachronic linguistics, where statistical power depends directly on sample size and where even subtle distributional patterns can carry theoretical significance.

Yet corpus growth has outpaced methodological adaptation. The bottleneck is not computational, as modern hardware can process billions of words in seconds, but human. For most syntactic and semantic research questions, corpus queries return vast quantities of data requiring manual classification before analysis can begin. For example, a linguist studying the dative alternation (*give the book to Mary* vs. *give Mary the book*; Bresnan et al., 2007) might sift through thousands of irrelevant tokens where *give* appears with unrelated senses or in idiomatic expressions. Meanwhile, a researcher investigating semantic change in polysemous modal verbs must distinguish, among other dimensions, between epistemic versus root uses (Coates, 1995), a task that standard NLP tools are hard-pressed to reliably achieve. These kinds of classifications demand linguistic expertise and cannot be fully automated with pattern-matching or off-the-shelf parsers. The result is a paradox: we have access to more data than ever before, but the cognitive and temporal costs of preparing that data for analysis limits our reach to small samples of these repositories.

The practical implications are substantial. Manual annotation of even 1,000 concordance lines can require days of focused work. Scaling to the tens of thousands of tokens necessary for robust quantitative analysis of synchronic or diachronic phenomena becomes prohibitively expensive in time or funding. The full analytical potential of century-spanning corpora remains underutilised, with researchers forced to trade sample size for historical coverage. Studies of rare constructions must accept sampling uncertainty or abandon questions altogether. Most fundamentally, the annotation bottleneck introduces the risk of systematic bias: researchers may tend to gravitate towards features that are easier to extract automatically, even when theoretically less interesting, while avoiding constructions that require extensive manual sorting despite their linguistic significance (see Marttinen Larsson, 2023 for a discussion of this issue in variationist sociolinguistics).

Against such a background, large language models (LLMs) represent a potentially transformative solution to this annotation challenge. Unlike traditional NLP tools that require domain-specific training data and programming expertise, conversational LLMs such as ChatGPT and Claude can be prompted in natural language to perform sophisticated classification tasks with minimal setup. Their emergent linguistic capabilities, though not without limitations (e.g. Cuskley et al., 2024; Leivada et al., 2024), include surprisingly refined grammatical and semantic judgments when properly prompted and supervised. Recent applications in linguistic annotation have shown promise across tasks ranging from sentiment analysis (Belal et al., 2023) to pragmatic phenomena (Yu et al., 2024) and most recently metaphor detection (Fuoli et al., 2026), suggesting that LLMs might serve as 'copilots for linguists' (Torrent et al., 2024) in corpus-based research.

In our previous work (Morin & Marttinen Larsson, 2025), we developed a proof-of-concept pipeline using Claude 3.5 Sonnet to annotate evaluative *consider* constructions through iterative, supervised training. Usage guides such as Merriam-Webster have suggested that 'the *consider* + *as* construction is becoming less and less common' (The Britannica Dictionary, n.d.), presumably being replaced by a zero complementizer (*We consider this approach Ø effective*) and, possibly, a *to be* complement (*We consider this approach* ***to be*** *effective*). Yet, no quantitative corpus study has tested this claim, likely because the data collection and annotation demands would be prohibitive using manual methods. Indeed, the lemma *consider*, which is polysemous and used both evaluatively (as above) and non-evaluatively (e.g. *She paused to consider her options*; *He considered that the proposal might fail*), appears in a myriad of variable forms, rendering data retrieval, cleaning, and annotation a tedious task for a human.

Morin and Marttinen Larsson (2025) established that conversational LLMs could achieve over 90% accuracy on this syntactically and semantically complex classification task in a held-out evaluation sample of 102 concordance lines, without requiring programming skills from the researcher and demonstrating both the feasibility and accessibility of LLM-assisted grammatical annotation. Despite the method's promise, however, it had not yet been tested against the scale of data required for robust diachronic research examining the variation and potential change undergone by evaluative *consider* (The Britannica Dictionary, n.d.). The question thus remained: could LLM tools extend from proof-of-concept to full corpus analysis spanning tens of thousands of tokens? Moreover, LLMs themselves have improved substantially since our initial study, with newer models offering enhanced accuracy and efficiency that might enable genuinely large-scale applications (e.g. Claude Sonnet 4.5, released in September 2025; Anthropic 2025; or GPT-5, released in August 2025; OpenAI, 2025).

Building on our preliminary work, this paper presents a fundamentally redesigned pipeline that achieves genuinely large-scale, automatic corpus annotation while maintaining, and indeed improving upon, the accuracy of our previous supervised approach. We conceived a streamlined, four-phase workflow that eliminates the need for continuous human supervision during the classification phase while maintaining an accuracy rate of 98%+ across multiple annotation tasks. Our pipeline consists of:

1. Comprehensive prompt engineering that encodes classification criteria, examples, edge cases, and output requirements
2. Pre-hoc evaluation on multiple independent random samples to validate prompt effectiveness before deployment
3. Automated batch processing via a LLM API, enabling the classification of tens of thousands of tokens without researcher intervention
4. Post-hoc validation using stratified random sampling, to verify classification reliability across subcorpora in a given dataset.

We demonstrate the pipeline's effectiveness through a large-scale case study of the English *consider* construction over two centuries. Using GPT-5 via the OpenAI API, we processed 143,933 concordance lines containing forms of *consider* from the entire Corpus of Historical American English (COHA; Davies, 2002) in under 60 hours — a task which, given a conservative estimate of 10 seconds per concordance line, would require at least 400 hours of manual annotation. The annotation involved two sequential tasks, each addressing a distinct methodological bottleneck. First, we distinguished 'evaluative' from 'non-evaluative' uses, a critical filtering step that separates genuine instances of the construction under investigation. This initial classification achieved 98%+ accuracy and reduced the dataset from 99,406 tokens to 44,527 evaluative instances, illustrating the efficiency of our approach for compiling datasets of minority variants. Second, we classified these evaluative instances by complement type — zero (*consider him intelligent*), *to be* (*consider him to be intelligent*), or *as* (*consider him as intelligent*) — achieving comparable accuracy. Post-hoc validation through stratified random sampling confirmed classification reliability across COHA's twenty decades (1820s–2010s) and five genres, showcasing the pipeline's robustness to historical and stylistic variation.

We argue that this methodological advance has momentous implications for what is practically achievable in corpus-based research. The pipeline requires minimal programming expertise beyond API setup, making sophisticated annotation accessible to the broader linguistic community for a wide range of tasks and features. By front-loading intellectual work into prompt engineering and validation rather than continuous supervision prone to human fatigue and error, our approach redefines the linguist's role in the design and execution of research projects. Crucially, we provide transparent documentation of computational costs: annotating 140k+ tokens across two classification tasks cost approximately $104 USD in API fees, which shows that large-scale LLM-assisted annotation is financially viable even for modestly funded studies, while saving weeks of full-time

work. The actual findings we make on the English *consider* construction are equally significant: a Bayesian multinomial generalized additive model fitted to 44,527 evaluative tokens reveals previously undocumented genre-specific trajectories of diachronic change, enabling us to advance new hypotheses about the relationship between register formality and competing pressures of morphosyntactic reduction and enhancement. Overall, the results validate both the technical accuracy of LLM-based annotation and its capacity to generate new linguistic insights in the field of constructional variation and change (Hoffmann & Trousdale, 2011).

The remainder of this paper is structured as follows. Section 2 provides background essentials on LLM-assisted corpus annotation approaches, the linguistic properties of the evaluative *consider* construction, and our previous pipeline for automatic grammatical annotation in large corpora. Section 3 details our new, four-phase methodological pipeline, including prompt design principles, validation strategies, LLM implementation via API use, and statistical approaches to post-hoc quality assurance. Section 4 presents our results across four dimensions: classification accuracy, computational efficiency, error analysis, and substantive linguistic findings about variation and change in the evaluative *consider* construction in COHA. Section 5 discusses methodological implications for corpus linguistics, addresses practical considerations including funding and data licensing, outlines future directions for LLM-assisted annotation research, and concludes by reflecting on the broader significance of this work for the field and the evolving relationship between human linguists and AI tools in corpus-based inquiry.[1]

## 2. LLM-assisted corpus annotation and the *consider* construction

This section situates our study within existing approaches to corpus annotation and the phenomenon it targets. We review the strengths and limitations of manual, rule-based, and supervised methods alongside the emerging role of LLMs (Section 2.1), introduce the evaluative *consider* construction and the complementizer variation that motivates our case study (Section 2.2), and summarise the supervised proof-of-concept pipeline on which the present work builds (Section 2.3).

1. Supplementary material referenced throughout this paper can be found at: https://osf.io/nuqcz/overview

## 2.1 LLM-assisted corpus annotation

The challenge of annotating corpus data has long been central to linguistic methodology (Gries & Berez, 2017). Traditional approaches fall into at least three categories with distinct limitations. First, manual annotation by human experts is the gold standard for nuanced linguistic judgment (Smith et al., 2008), but is labour-intensive and difficult to scale beyond thousands of tokens. Second, rule-based automatic systems using pattern-matching offer speed (Hovy & Lavid, 2010), but lack flexibility for semantically rich or context-sensitive linguistic structures. Third, supervised machine learning approaches provide particularly accurate and fine-grained results (Fonteyn et al., 2025), but require substantial training data and technical expertise, creating barriers for many linguists. Moreover, these approaches are typically language- and phenomenon-specific, potentially encouraging further focus on case studies with well-established annotation models.

By contrast, recent developments in LLM technologies have introduced new possibilities to overcome these obstacles. Unlike earlier approaches to automation, LLMs can engage in highly subtle linguistic analyses based on prompts written in natural language, leveraging increasingly voluminous pre-training datasets in successive model generations (Kaplan et al., 2020). Early work deploying LLMs for corpus linguistic research has demonstrated promising results across diverse annotation types. Belal et al. (2023) showed ChatGPT's effectiveness for sentiment analysis; Yu et al. (2024) achieved reasonable accuracy on context-sensitive pragmatic phenomena such as apology expressions; Fuoli et al. (2026) demonstrated that state-of-the-art models can achieve median F1 scores of 0.79 on metaphor identification using prompt engineering approaches; and Ostyakova et al. (2025) deployed a LLM method to perform discourse and speech act annotation that matched or even surpassed human evaluators.

The specific annotation problem motivating the present study lies in the domain of syntactic variation and change (Morin & Marttinen Larsson, 2025). While research in this field can often take advantage of automatic parsers to identify formal structural patterns, the semantic dimensions of grammatical constructions pose a more fundamental challenge. Since constructions are theorised as form-meaning pairings (Leclercq & Morin 2025), understanding their distribution and evolution requires distinguishing not only syntactic variants but also the wider contexts that make up their conceptual content (Geeraerts et al., 2023). Such distinctions demand fine-grained judgments of the kind that has generally required human expertise, creating a clear impediment to large-scale quantitative research. We illustrate the issue by turning to the case study at the heart of this paper: the evaluative *consider* construction in English.

## 2.2 A case study of complementizer variation in evaluative *consider* constructions

The English verb *consider* participates in multiple grammatical constructions with distinct meanings. Among these is an evaluative use (Jacques, 2023) where the verb introduces a characterisation or judgment about an entity. In its evaluative use, *consider* takes an object (the entity being evaluated) and a predicative complement (the evaluation itself), as illustrated in Example (1):

(1) a. They consider him Ø intelligent. (zero complement)
b. They consider him *to be* intelligent. (*to be* complement)
c. They consider him *as* intelligent. (*as* complement)

The three variants in Example (1) appear to be semantically 'equivalent' (Leclercq & Morin, 2023), differing only in how the complement is formally realised: directly with no additional marking (1a), through the infinitival copula *to be* (1b), or via the preposition *as* (1c). This variation extends to the passive voice (*He is considered intelligent/to be intelligent/as intelligent*), multiplying the surface realisations that a researcher studying this construction would need to track in a corpus. Intervening materials of varying length readily occur between the verb and the complement, sometimes spanning entire phrases and clauses (e.g. *The patient was considered by the doctors and nurses* ***to be*** *particularly demanding*; *We consider matters pertaining to legislation and the penal code* ***as*** *urgent; I consider him, as far as our job is concerned,* ***Ø*** *a very competent colleague*). This discontinuity means that simple pattern-matching or n-gram searches cannot reliably capture the full range of the construction's usage, challenging the prospects of exhaustive corpus extraction.

In addition, although the evaluative *consider* construction is straightforward to identify on formal and functional grounds, it coexists with another highly frequent use of the verb that emerges when searching the lemma in large datasets. In this non-evaluative use, *consider* functions as a cognition verb meaning 'contemplate', 'take into account', or 'think about', as illustrated in Example (2). Crucially, these non-evaluative uses lack the object complement structure that characterises evaluative constructions: they feature simple direct objects (2a), *that*-clauses (2b), or *for*-phrases indicating purpose (2c), but no predicative element that evaluates or characterises the object.[2]

**2.** The monoclausal evaluative *consider* construction investigated here differs fundamentally from multiclausal *[consider + that-clause]* patterns in both structure and function: the former directly characterises an entity through a predicative complement, while the latter expresses deliberation about propositional content. Although Jacques (2023: 178–179) groups both under the label 'estimative', we restrict our analysis to monoclausal patterns and treat *that*-clauses as cognition verb uses. See Morin and Marttinen Larsson (2025) for detailed justification.

(2) a. Please *consider* all options before deciding.
b. The committee *considered* that the proposal had merit.
c. She was *considered* for the position but not hired.

In Morin and Marttinen Larsson (2025), an initial search of *consider* tokens in the EnTenTen21 web corpus yielded as many as 18 million tokens. A 200-token random sample revealed that only 11.5% qualified as evaluative constructions, while the remaining 88.5% comprised non-evaluative uses of 'cognitive' *consider* (as in Example (2)). Even with sophisticated corpus query syntax, researchers would need to manually inspect hundreds of thousands of concordance lines, simply to obtain datasets of sufficient size for analysing the evaluative construction's formal variants, let alone their diachronic trajectories.

Overall, these practical constraints motivated us to develop a methodological pipeline for automated annotation of *consider* using LLMs, with a view to making it replicable and adaptable to diverse linguistic classification tasks, constructions, and languages.

### 2.3 An iterative, supervised training pipeline for LLM-based corpus annotation

In our previous work (Morin & Marttinen Larsson, 2025), we developed a proof-of-concept pipeline for LLM-assisted grammatical annotation that demonstrated the feasibility of using conversational AI models to classify evaluative *consider* constructions with high accuracy. The pipeline, showcased in Figure 1, was designed to be accessible to linguists without programming expertise, relying instead on prompt engineering and iterative training through a chat interface. Using Claude 3.5 Sonnet, we achieved over 90% accuracy in distinguishing evaluative from non-evaluative uses of *consider* through a structured three-phase process.

The approach designed by Morin and Marttinen Larsson (2025) was fundamentally iterative and largely supervised, relying on a back-and-forth training process during which the LLM received corrective feedback to progressively refine its understanding of the classification criteria. This process was constrained to the chat environment available through Claude's interface. Additionally, our previous approach had a narrower scope, addressing only the distinction between non-evaluative and evaluative uses rather than also classifying the complement forms (*zero*, *to be*, *as*) of evaluative *consider*. Lastly, while this preliminary approach validated the core concept of LLM-assisted annotation, it relied on continuous human supervision throughout the classification phase, and had not yet been tested at a larger scale.

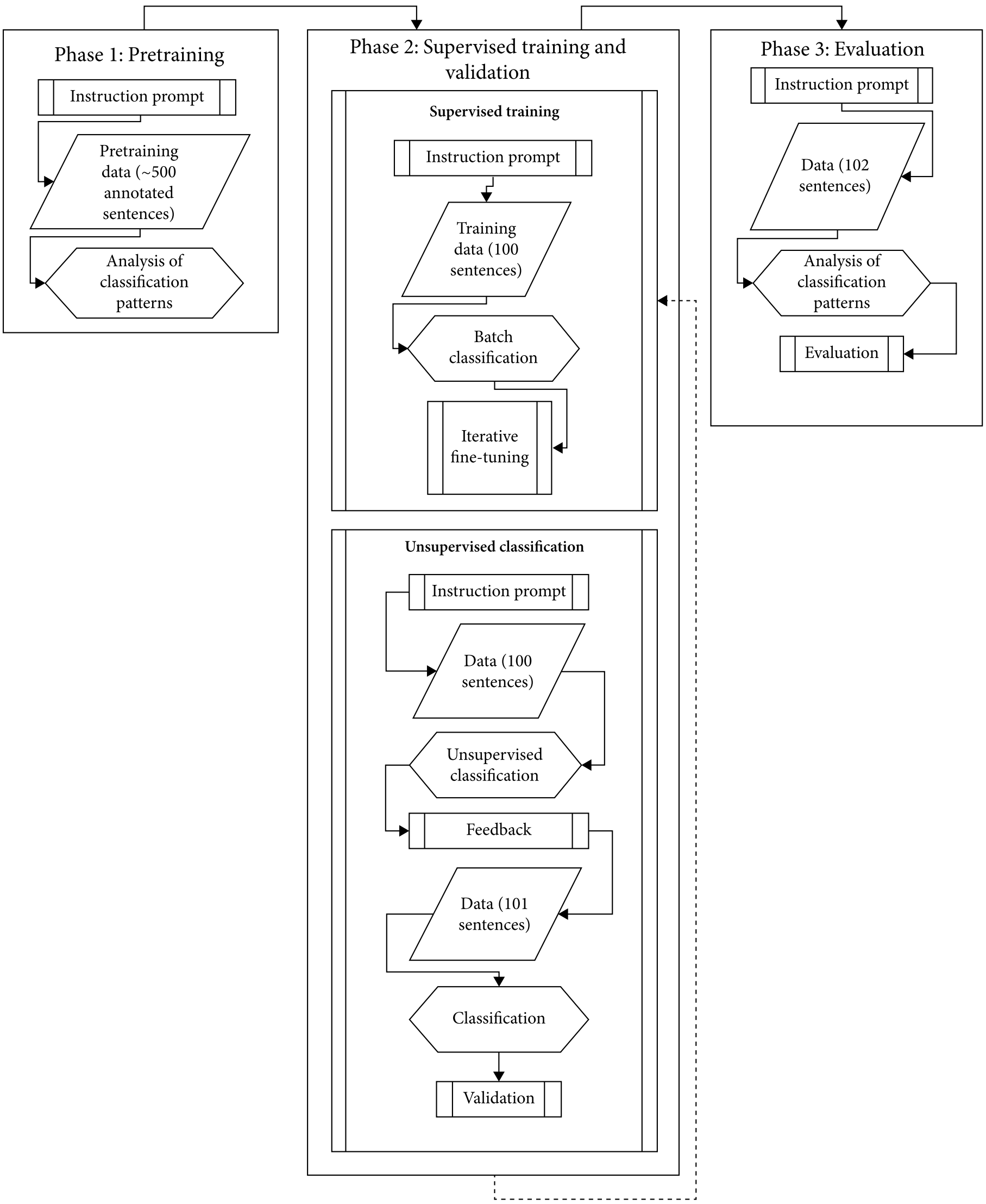

**Figure 1.** Morin and Marttinen Larsson's (2025) pipeline for LLM-assisted annotation of linguistic data, based on the case study of *consider* constructions

The proof-of-concept results demonstrated that LLM-assisted annotation could achieve expert-level reliability on complex grammatical classification tasks. Building on these foundations, the present study introduces a fundamentally redesigned pipeline that preserves the core principles of effective prompt engineering while enabling large-scale, automatic annotation with enhanced accuracy and no pre-training, as detailed in the next section.

## 3. The annotation pipeline and its implementation in COHA

This section presents our methodological contribution in two parts. We first describe the four-phase pipeline for automatic corpus annotation in general terms, so that it can be adapted to other tasks and languages (Section 3.1), and then detail its implementation in our diachronic case study of evaluative *consider* in COHA, covering data extraction, LLM selection, and task-specific prompt engineering (Section 3.2).

### 3.1 A four-phase pipeline for automatic corpus annotation

Based on the tenets of Section 2.3, we now present a new pipeline optimised for large-scale, automatic corpus annotation. The pipeline consists of four sequential phases: (1) comprehensive prompt engineering, (2) pre-hoc evaluation, (3) automated batch processing, and (4) post-hoc validation. This workflow front-loads the intellectual labour into prompt design and validation, enabling the LLM to process tens of thousands of tokens autonomously once the system has been validated. Figure 2 illustrates the complete workflow.

The first phase involves engineering a detailed instruction prompt that encodes all classification criteria, representative examples, edge case handling, and output formatting requirements. Drawing on the guidelines of Section 2.3, effective prompts for grammatical annotation should include: (a) explicit task instructions that specify the classification goal and desired accuracy threshold; (b) illustrative examples covering the full range of target constructions, including prototypical instances and boundary cases; (c) structured formatting using XML tags (e.g. <instructions>, <examples>, <thinking>) to help the LLM parse different components of the prompt; (d) Chain of Thought prompting that instructs the model to reason step-by-step through its classification decisions; and (e) precise output format specifications that ensure consistency across large batches of data. Crucially, the prompt must anticipate potential sources of misclassification by explicitly addressing ambiguous patterns, structurally complex examples, and formally similar but functionally distinct constructions. The goal of this phase is to create a self-contained instruction set that enables accurate classification without real-time human intervention.

Before deploying the classification prompt on the full corpus, Phase 2 validates its effectiveness through evaluation on multiple independent random samples drawn from the target dataset. We recommend conducting at least three independent evaluations on samples of a size appropriate to the classification task and LLM capabilities. We also recommend evaluating multiple batch sizes (e.g. 50, 100, 200 concordance lines) on independent random samples, calculating accuracy for each, and examining whether errors cluster toward the end of batches, which would indicate attention degradation. The optimal batch size is the largest that maintains target accuracy (95%+), with errors distributed randomly rather than concentrated in later portions. In our implementation, 100-concordance line batches maintained 98%+ accuracy, while preliminary testing at 200 concordance lines revealed persistent performance degradation after c. 100 concordance lines. For each evaluation sample, the researcher first manually annotates the concordance lines to establish ground truth. Meanwhile, the LLM receives the raw concordance lines without any annotations, and classifies them independently. The LLM's classifications are compared against the researcher's ground truth to calculate accuracy metrics (see Section 3.2).

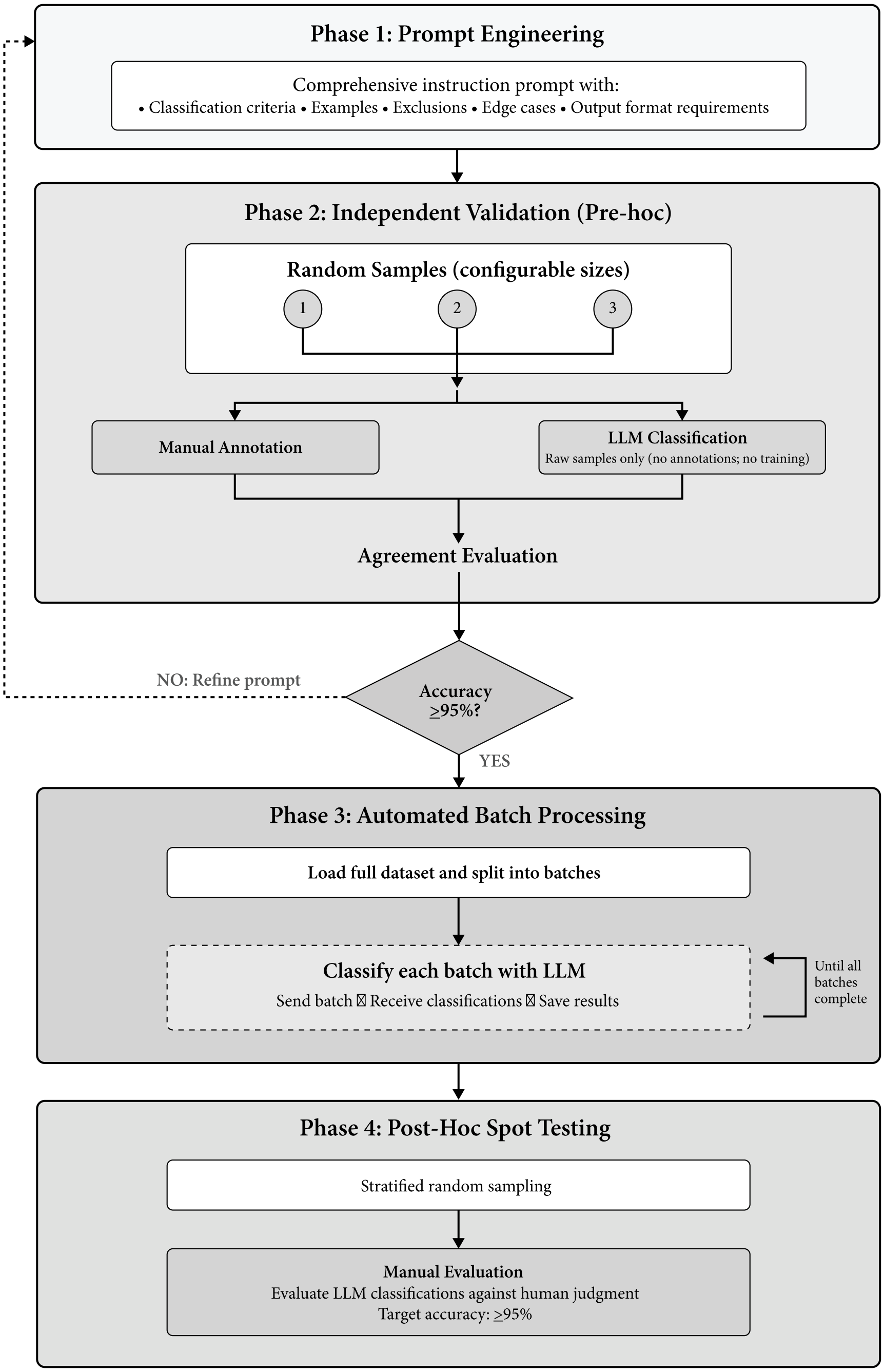

**Figure 2.** The four-phase pipeline for automated, large-scale corpus annotation using LLMs (Morin & Marttinen Larsson, 2026)

The accuracy threshold required to proceed to Phase 3 depends on the research application, but for most corpus-based studies of grammatical variation, we recommend that it should exceed 95% observed accuracy for each individual sample. This stringent criterion ensures that even accounting for sampling uncertainty, the LLM's true classification accuracy is likely to be sufficiently high for large-scale deployment. If pre-hoc evaluation reveals systematic misclassifications, the prompt should be iteratively refined before re-evaluation. Researchers should examine patterns in the errors (e.g. identifying whether misclassifications cluster around particular structural configurations, ambiguous examples, or specific lexical contexts), and revise the prompt to address these issues explicitly. This may involve adding clarifying instructions, providing additional examples of problematic cases, or restructuring the decision criteria. The pre-hoc evaluation cycle continues until the accuracy threshold is met across all independent samples.

Once the prompt has been validated, Phase 3 involves deploying it at scale via a LLM API to process the full corpus dataset. Unlike conversational interfaces that limit input size and require manual interaction, API access enables programmatic submission of large batches of concordance lines for classification without human intervention during the processing phase. Concordance lines should be processed in batches matching the sample size validated in Phase 2, ensuring that the LLM maintains the same level of attention and classification quality observed during pre-hoc evaluation. Using a script, each batch is iteratively submitted to the API along with the instruction prompt, receives the classifications, and aggregates the results. While the specific implementation depends on the chosen API and programming environment, the basic workflow involves: (a) dividing the corpus data into appropriately sized batches; (b) formatting each batch according to API requirements; (c) submitting batches sequentially or in parallel (depending on rate limits); (d) capturing and storing the LLM's output for each batch; and (e) compiling the classifications into a unified dataset for analysis. We provide a Python implementation script in the supplementary material that can be adapted to different APIs and research contexts; equivalent functionality can be achieved in R or other programming languages. Specific API configuration details, including rate limiting and cost management, are discussed in Sections 3.2.2 and 3.2.4 in the context of our case study implementation.

The final phase involves verifying classification reliability across the complete annotated dataset through stratified random sampling. For example, to ensure that validation accounted for potential variation in corpus composition, our samples were stratified by decade-genre combinations, with proportional allocation based on the distribution of extracted tokens for the phenomenon of interest (see supplementary material for the code scripts). This stratification strategy ensures that the validation sample mirrors the subcorpora dimensions of the entire dataset. In our implementation, this yielded approximately 85 strata (corresponding to non-empty decade-genre cells across COHA's 20 decades and five genres), with samples drawn randomly from each stratum across all batches. If, however, the researcher's dataset does not contain distinct subcorpora, we recommend stratifying the random sampling procedure by individual batches. In our case, this would have led to (995 + 446 =) 1,441 strata to sample from.

The validation sample size should be calculated to achieve adequate statistical precision, typically targeting a margin of error of ±1% at 95% confidence. Using the standard formula for binomial proportion confidence intervals with finite population correction, researchers can determine the required sample size:

$$n = n_0 / (1 + (n_0 - 1) / N) \text{ where } n_0 = (z^2 p(1 - p))/ME^2$$

Setting parameters at a 95% confidence level ($z = 1.96$), expected accuracy based on pre-hoc evaluation (e.g. $p = 0.96$ for 96% accuracy), and desired margin of error ($ME = 0.01$ for ±1%), we can calculate the target sample size given their total dataset size (N). For large corpora (e.g. from 50,000 to 150,000 concordance lines), this typically yields validation samples representing 1–3% of the total dataset. These validation concordance lines should be manually annotated by the researcher and compared against the LLM's classifications to calculate accuracy metrics across strata.

If post-hoc validation reveals that the final accuracy falls below the pre-hoc threshold, a judgment call must be made based on the specific research context. First, the error patterns should be examined closely: are misclassifications random, or do they reflect systematic issues with particular constructions or strata in the dataset? Second, researchers should consider whether the observed accuracy (e.g. 90–95%) is adequate for their research question and acceptable within their field's standards, consulting relevant methodological literature for guidance.

In some cases, slightly lower accuracy may be sufficient, particularly if errors are randomly distributed. However, if accuracy is inadequate, the remedy depends on the error structure. Systematic errors confined to an identifiable subset — a stratum, or lines matching a detectable pattern — can be isolated, reprocessed under a refined prompt, and re-validated. Random errors cannot be located individually; the researcher must then either accept and transparently report the residual noise (see Section 4.2) or reprocess the full dataset under an improved prompt or model. The key principle is that post-hoc validation provides transparency about classification reliability, enabling the linguist to make informed decisions about data quality and to report limitations appropriately.

## 3.2 Case study implementation: Evaluative *consider* in COHA

The remainder of this section applies the pipeline described in Section 3.1 to our case study of the evaluative *consider* construction in COHA. We present the corpus, the extraction procedure and the two successive classification tasks (Section 3.2.1), the choice of GPT-5 and the processing run with its duration and cost (Section 3.2.2), the two task-specific prompts (Section 3.2.3), and the API configuration and data-sharing settings under which all calls were made (Section 3.2.4). The two evaluation phases, pre-hoc and post-hoc, are reported alongside the results in Section 4.1.

### 3.2.1 *Data*

Our data comes from the Corpus of Historical American English (COHA; Davies, 2002), a balanced diachronic corpus containing approximately 556 million words over 20 decades from the 1820s through the 2010s. The corpus is split across five genres: academic prose, fiction, magazines, newspapers, and transcripts from TV and movies. We obtained a private institutional license to analyse the corpus through Université Paris-Cité, which permitted computational processing of the data subject to specific usage restrictions. These included a requirement that all LLM API processing be conducted with data sharing disabled, to prevent COHA materials from entering model training datasets (we return to data licensing considerations in Section 5). The corpus creator, Mark Davies, granted explicit permission for this research application via personal communication (October 2025).

To extract instances of the *consider* construction, we wrote a Python script that searched all COHA text files for inflected forms of the lemma CONSIDER: *consider*, *considers*, *considering*, and *considered*. The script explicitly excluded lexical items containing the string "consider" but belonging to different lemmas, such as *reconsider*, *inconsiderate*, *considerable*, *considerably*, *considerate*, *considerately*, *consideration*, and *reconsideration*. For each matching token, the script extracted a context window of 15 words before and 15 words after the target form (roughly similar to the default UI context windows of English-corpora.org), providing sufficient context for argument structure classification while maintaining manageable batch sizes for API processing.[3] Crucially, the script preserved metadata for each instance, recording the decade and genre of origin to enable subsequent stratified validation and diachronic analysis. This initial extraction yielded 99,406 instances of CONSIDER distributed across the entire corpus.

The annotation task involved two sequential classification steps, each addressing a distinct aspect of the construction's structure and variation. First, we classified all 99,406 instances as either evaluative or non-evaluative uses of *consider* (as described in Section 2.2), reducing the dataset to 44,527 evaluative tokens. Second, we classified these evaluative instances according to their complement variant: zero, *to be*, or *as*. The combined annotation effort thus processed 143,933 concordance lines (99,406 + 44,527) across both tasks.

### 3.2.2 *LLM selection and implementation*

For this study, we selected GPT-5 (OpenAI, 2025) accessed via the OpenAI API, rather than continuing with Claude 3.5 Sonnet used in our previous work. This choice was motivated by two primary considerations. First, preliminary API testing on validation samples revealed that GPT-5 achieved substantially higher accuracy on the *consider* classification tasks compared to both Claude 3.5 Sonnet and the recently released Claude Sonnet 4.5. Second, GPT-5 offered substantially lower API costs: at the time of processing, GPT-5 was priced at $1.25 per million input tokens and $10.00 per million output tokens, compared to Claude Sonnet 4.5's pricing of $3.00–$6.00 per million input tokens and $15.00–$22.50 per million output tokens, depending on prompt length. Given the scale of our

3. The relationship between context window size and classification accuracy is an open methodological question for LLM-assisted annotation. Larger windows provide more contextual information but increase token consumption and API costs proportionally, with environmental implications (see Section 5); smaller windows are more efficient but risk truncating relevant syntactic material. For argument structure constructions (Goldberg, 1995) such as *consider*, whose core elements (verb, object, and complement in evaluative uses) typically occur in close proximity to the target lemma, a ±15-word window proved empirically sufficient. To verify this, both authors independently annotated a random sample of 300 concordance lines stratified by token distribution in the subcorpora, assessing whether the verb, object, and (where applicable) predicative complement were identifiable within each concordance line. Agreement was 98.0% (294/300 items; PABAK = 0.960, indicating near-perfect reliability). We report PABAK rather than Cohen's $\kappa$ (Byrt et al., 1993) because the latter was not interpretable due to extreme class imbalance (only six negative cases across both raters). Future work might systematically compare window sizes across different construction types to establish general guidelines for this parameter.

annotation task (processing nearly 144,000 concordance lines across two sequential classification steps), these cost differences translated to meaningful budgetary implications for the research project.

We implemented the annotation workflow using OpenAI's API with the model string "gpt-5". Batches were processed one after the other, with each batch containing 100 concordance lines as validated during pre-hoc evaluation (Phase 2). We set a maximum token limit of 20,000 tokens per API call to accommodate both the instruction prompt and the expected output classifications, while temperature parameters (controlling output randomness) remained at their default values. Overall, we encountered no rate limiting issues during processing, as our workflow remained well within OpenAI's usage tier thresholds.

The two successive annotation tasks each required separate API processing runs. For Task 1, we processed all 99,406 extracted instances in 995 batches of 100 concordance lines (with the final batch containing six lines). This classification consumed approximately 12.45 million tokens, and required approximately 44.4 hours of processing time. Task 2 processed the resulting 44,527 evaluative instances in 446 batches of 100 concordance lines (with the final batch containing 27 lines). This second task consumed approximately 4.96 million tokens, and required approximately 14.2 hours of processing time.

Combined, the two tasks thus processed 143,933 concordance classifications in approximately 58.6 hours, at a total cost of approximately $104 USD in API fees (calculated as $75.38 for Task 1 and $28.21 for Task 2, excluding $3.51 in preliminary testing costs).

### 3.2.3 *Task-specific prompt engineering*

Following the prompt engineering principles from Phase 1 of our pipeline (Section 3.1), we developed two task-specific instruction prompts: one for distinguishing evaluative from non-evaluative uses of *consider*, and one for classifying evaluative instances by complement variant. Both prompts incorporated explicit task instructions, illustrative examples, XML structuring tags, Chain of Thought reasoning directives, and precise output formatting requirements.

The first prompt instructed GPT-5 to perform binary classification of *consider* tokens, targeting a minimum accuracy of 95%. The prompt opened with explicit exclusions of lexical items such as *considerable*, *consideration*, and *reconsider*, then systematically described evaluative uses as featuring subject or object complements that characterise or evaluate an entity (e.g. *They considered Lena the best swimmer*; *Brooklyn is not considered part of Manhattan*). It explained all three complement variants – *as*, *to be*, and zero – emphasizing that the zero variant (direct complement without additional marking) is equally evaluative despite lacking overt morphosyntactic signals. A critical component was what we called the 'insertion test', which instructed GPT-5 to test whether *to be* or *as* could be inserted after the object while maintaining grammatical acceptability and meaning (e.g. *I consider him my best friend → I consider him [to be] my best friend* ✓ evaluative; *They considered the proposal → They considered the proposal [to be]* ? ✗ non-evaluative). The prompt also addressed edge cases that preliminary testing revealed as sources of misclassification, including adverbial participles (e.g. *Historically considered, the treaty was a turning point*), fronted adjectives (e.g. *However trivial the board may consider it, the bug must be fixed*), and *as*-clause attribution patterns (e.g. *as it is considered by rivals, the spoiler faction*).

The second prompt classified the 44,527 evaluative instances according to their complement variant: zero (*He is considered intelligent*), *to_be* (*He is considered to be intelligent*), or *as* (*He is considered as intelligent*). This prompt opened by clarifying that all concordance lines were already classified as evaluative, narrowing the task to variant identification. It provided systematic examples across active and passive voice constructions and devoted substantial attention to edge cases, such as *as* appearing elsewhere in the concordance line without introducing the complement (e.g. *As they see it, this is considered a problem*), ellipsis and interruption patterns, and the distinction between copular *to be* and other infinitival structures (e.g. *They consider him good to go*). Similarly to Task 1, it incorporated Chain of Thought instructions directing GPT-5 through a stepwise diagnostic process.

Both prompts concluded with strict output formatting requirements specifying that GPT-5 must produce exactly one numbered classification per concordance line with no additional text, enabling automated parsing during batch processing. The complete prompts for both classification tasks are provided as standalone text files in the supplementary material, and are also included within the API implementation scripts.

### 3.2.4 *API configuration details*

All API calls used the gpt-5-2025-08-07 model via OpenAI's responses.create endpoint, with processing completed between October 1–7, 2025. Run dates are verifiable via usage logs downloadable from the OpenAI API platform. Each request consisted of a single user-role message containing the full prompt and 100 numbered concordance lines; no system or developer message was used, with all instructions embedded directly in the prompt. Sampling parameters (temperature and top_p) were left at API defaults of 1.0, and max_output_tokens was set to 20,000 to accommodate Chain of Thought reasoning. Failed requests were retried up to three times with 3-second intervals. Data sharing was disabled via the OpenAI platform's organisation settings (Organisation → Data controls → Sharing settings), where three options are available: (i) 'Enable sharing of model feedback from the Platform', (ii) 'Share evaluation and fine-tuning data with OpenAI', and (iii) 'Share inputs and outputs with OpenAI'. All three were set to 'Disabled' for this project, ensuring that neither the corpus data nor the model outputs were used to train OpenAI's models.

A reviewer noted that temperature = 0 is often recommended for classification to reduce output variability; we deliberately retained the default to establish baseline performance under the out-of-the-box settings a novice user would encounter. In any case, recent work shows that even temperature = 0 is not fully deterministic: Atıl et al. (2025: 135) report accuracy swings of up to 15% across identical runs, concluding that no LLM reliably delivers identical outputs regardless of task. Our pipeline nonetheless reached 98–99% accuracy under default settings, achieving robustness of the kind that the authors attribute to short, constrained outputs.

## 4. Results: Pipeline performance and diachronic findings

This section reports our findings in three steps. We first establish the pipeline's classification accuracy and computational efficiency (Section 4.1), then scrutinise the small number of misclassifications and their distribution across decades and genres (Section 4.2), and finally present the substantive diachronic findings on variation and change in the evaluative *consider* construction that our annotated dataset makes possible (Section 4.3).

### 4.1 Classification accuracy and computational efficiency

Following Phase 2 of our pipeline, we first validated the classification prompts on three independent random samples of 100 concordance lines each, drawn from the full COHA dataset using different random seeds. Table 1 presents the evaluation metrics for both tasks. For Task 1 (evaluative versus non-evaluative classification), GPT-5 achieved accuracies of 98%, 98%, and 100% across the three samples, with an average accuracy of 98.67%. For Task 2 (variant classification), GPT-5 achieved accuracies of 98%, 99%, and 99% across the three samples, with an average accuracy of 98.67%. These results clearly validate the prompts for large-scale deployment by exceeding our pre-established threshold, which required a value exceeding 95%.

Following Phase 4 of our pipeline, we also conducted post-hoc validation through stratified random sampling. Based on the sample size calculation formula introduced in Section 3.1, we determined target samples of 1,454 concordance lines for Task 1 (1.5% of 99,406 total) and 1,428 concordance lines for Task 2 (3.2% of 44,527 total). To ensure the validation sample mirrored the composition of the classified dataset, we stratified sampling across all decade-genre combinations using proportional allocation based on the distribution of extracted *consider* tokens: each stratum (e.g. 1920s fiction, 1850s newspapers) received a number of validation concordance lines proportional to its share of the extracted dataset. Manual verification of these stratified samples revealed accuracies of 99.04% for Task 1 (1,440/1,454 correct, with a minimum of 91.30%, median of 100%, and maximum of 100% across strata) and 99.16% for Task 2 (1,416/1,428 correct, with a minimum of 87.50%, a median of 100% and a maximum of 100%), confirming that the pipeline maintained high classification reliability across the complete dataset, and demonstrating that accuracy remained stable when deployed at scale.

**Table 1.** Pre-hoc evaluation metrics for classification tasks. Task 2 metrics are macro-averaged for multi-class classification

| Task | Sample 1 (*N*= 100) | Sample 2 (*N*= 100) | Sample 3 (*N*= 100) | Average | Precision | Recall | F1 Score | MCC |
|---|---|---|---|---|---|---|---|---|
| Task 1 | 98% | 98% | 100% | 98.67% | 98.12% | 99.3% | 98.72% | 0.9735 |
| Task 2 | 98% | 99% | 99% | 98.67% | 97.20% | 98.12% | 97.50% | 0.9676 |

As foreshadowed in Section 3.2.2, the automated batch processing exhibited substantial efficiency gains compared to what manual annotation would have likely yielded. Task 1 processed 99,406 concordance lines in 995 batches with an average processing time of 158.6 seconds per batch (range: 20.3–410.9 seconds), while Task 2 processed 44,527 evaluative concordance lines in 446 batches with an average processing time of 112.7 seconds per batch (range: 53.8–236.6 seconds). We return to the topic of efficiency in Section 5.

## 4.2 Error analysis

Although the pipeline achieved very high accuracy for both tasks, a systematic examination of the small number of misclassifications can reveal informative patterns about the model's decision boundaries and potential edge cases. We recommend that researchers using LLM-assisted annotation pipelines conduct similar error analyses, even when overall accuracy is high, as such analyses can (i) identify systematic biases that might affect downstream conclusions, (ii) inform prompt refinement for future applications, and (iii) provide qualitative insight into the linguistic phenomena that challenge automated classification.

Firstly, Table 2 presents the confusion matrix for Task 1. Of the 14 errors out of 1,454 concordance lines, 11 were false positives (non-evaluative instances misclassified as evaluative) and three were false negatives (evaluative instances mis-classified as non-evaluative). This asymmetry indicates that the model erred very slightly towards over-classification as evaluative.

**Table 2.** Task 1 confusion matrix (Precision: 98.30%; Recall: 99.53%; F1: 98.91%)

| | **Predicted: Evaluative** | **Predicted: Non-evaluative** | **Total** |
|---|---|---|---|
| Actual: Evaluative | 637 (True positive) | 3 (False negative) | 640 |
| Actual: Non-evaluative | 11 (False positive) | 803 (True negative) | 814 |
| **Total** | **648** | **806** | **1,454** |

Qualitative examination of the false positives revealed a few ways in which the model misinterpreted a surface structure containing *consider* as involving an evaluative predicative complement (Examples (3a)–(c)). In Example (3a), *alike* is not a postposed modifying adjective but an adverb, thus functioning as an adverbial; in Example (3b), *as though* is a conjunction that introduces a clause, not a noun phrase; and in Example (3c), *at rest* is a descriptive complement rather than an evaluative one.

(3) a. I know people *consider* all mediums alike.
  b. The mind shall be *considered* and treated by the great mass of mankind as though it were of equal consequence.
  c. If, on the other hand, we *consider* the warm body at rest in the second case mentioned above …

Secondly, Table 3 presents the confusion matrix for Task 2, which totals 12 mistakes for 1,428 concordance lines (see Examples (4a)–(c)). This matrix includes an additional row for N/A cases: occasional instances (nine) that should not have entered Task 2 because they were non-evaluative uses misclassified in Task 1. Meanwhile, only three errors were true variant misclassifications within the evaluative subset.

**Table 3.** Task 2 confusion matrix (Per-class F1: Zero = 0.99; to be = 0.98; as = 1; weighted average F1 = 0.99)

| | **Predicted: *zero*** | **Predicted: *to be*** | **Predicted: *as*** | **Total** |
|---|---|---|---|---|
| Actual: *zero* | 1,060 | 2 | 0 | 1,062 |
| Actual: *to be* | 1 | 110 | 0 | 111 |
| Actual: *as* | 0 | 0 | 246 | 246 |
| Actual: N/A | 8 | 1 | 0 | 9 |
| **Total** | **1,069** | **113** | **246** | **1,428** |

(4) a. There are lesions of the will that are no more to be *considered* subject to moral condemnation than a strain of the spinal column. [misclassified as *to be* variant]

b. The first phase may be *considered* to terminate about 1470 or thereabouts. [misclassified as zero variant]

c. Although I have never seriously *considered* a romantic relationship... [N/A non-evaluative]

Notably, the *as* variant achieved perfect classification ($F1 = 1$), likely because the explicit lexical marker *as* provides a clear signal when the status as evaluative is ascertained. Meanwhile, the *to be* variant had the lowest (though still excellent) F1 of 0.9821, with two instances misclassified as zero (Examples (4a) and (4b)).

To assess whether errors could systematically bias our diachronic or register-based conclusions (see Section 4.3), we examined their distribution across decades and registers. For Task 1, the 14 errors were distributed across four registers and 10 decades, with error rates ranging from 0.00% (TV/movies) to 1.52% (academic). For Task 2, the 12 errors were distributed across all five registers and eight decades, with error rates ranging from 0.47% (magazine) to 2.08% (TV/movies, though this higher rate reflects the small stratum size of $n = 48$). Crucially, errors showed no temporal clustering, appearing in decades spanning the 1820s to the 2010s. This dispersion indicates that classification errors will have exceedingly little impact on the linguistic patterns that may emerge from the data.

Finally, we calculated the maximum change in variant proportions that would result from correcting all errors. This involved removing the nine upstream N/A cases (reducing the denominator from 1,428 to 1,419) and reclassifying the three true variant errors. Table 4 presents the results.

**Table 4.** Proportion shift under error correction

| **Variant** | **Current** | **Corrected** | **Shift** |
|---|---|---|---|
| *zero* | 74.86% | 74.84% | −0.02 pp |
| *to be* | 7.91% | 7.82% | −0.09 pp |
| *as* | 17.23% | 17.34% | +0.11 pp |

The maximum proportion shift is 0.11 percentage points: well below any threshold that could affect linguistic explorations of the data. Even under complete reclassification of all errors, the relative ordering and approximate magnitudes of variant frequencies remain unchanged.

## 4.3 Linguistic findings

In this subsection, we address two research questions enabled by our automatically annotated, 44,500+ data points:[4] (i) what diachronic changes has the evaluative *consider* construction undergone between the 1820s and 2010s?; and (ii) what genres have these changes taken place in?

To begin with, we note that, panchronically, the zero variant is the most frequent variant (*N*=33,198, 74.55%), followed by the *as*-variant (*N*=7809, 17.53%) and, finally, the *to be* variant (*N*=3520, 7.9%). Figure 3 presents the distribution of *consider* variants as proportions across COHA's twenty decades. Notably, the data confirms The Britannica Dictionary's suggestion that the *as*-complement has declined substantially over time. However, our comprehensive diachronic evidence reveals a more complex trajectory than previously documented. By the 1820s, the *zero* and *as* variants appear at roughly equal frequencies, but the distributional pattern suggests that the *zero* complement had already been replacing *as* for some time before COHA's temporal coverage begins. There are thus reasons to believe that the change was already well advanced by the early nineteenth century.

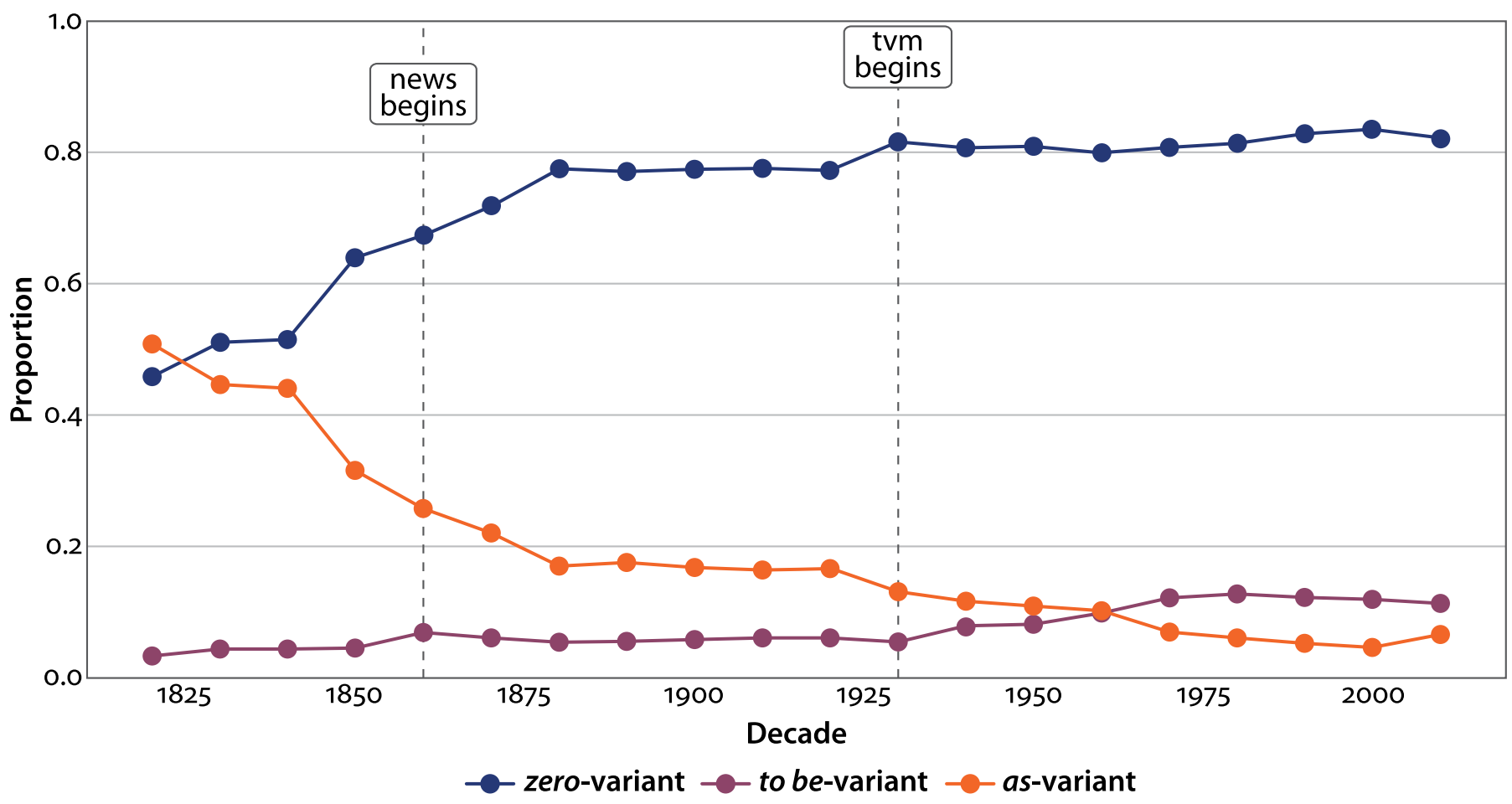

**Figure 3.** Diachronic evolution of *consider* variants over time (all genres combined, News = newspapers sub-corpus, tvm = TV/movie sub-corpus)

4. A heatmap summarising the distribution of evaluative *consider* tokens by decade and genre (44,527 total) is provided in the supplementary material to contextualise the analyses that follow.

In order to ascertain the diachronic trajectory undergone by the three *consider* variants and their diffusion across usage contexts, we fitted a Bayesian multinomial generalized additive model (GAM) through the *brms* package (Bürkner, 2017) in R, using CmdStan as the computational backend (Carpenter et al., 2017). The response variable was the three-level complementation variant (zero, *as*, *to be*), modelled with a categorical likelihood and using zero – the most frequent level – as the reference category. The model was specified as follows:

variant ~ genre + s(decade_c, by = genre, $k=8$), family = categorical(refcat = "zero")

The formula included genre as a main effect (estimating baseline differences among the five genres on the log-odds scale) and a thin-plate regression spline over 'decade' for each genre, specified via the 'by = genre' term. This structure allowed each genre to follow its own non-linear trajectory over time without assuming any particular form. The number of basis functions was set to $k=8$. This provides sufficient flexibility for the 20 decades of data while guarding against overfitting, particularly for the TV/movie genre which only consisted of nine decades of observations.

The 'decade' variable was centred on 1920, which is near the middle of the observed timespan, and scaled to decade units to improve numerical stability and ensure that the model intercept corresponded to a meaningful timepoint near the centre of the observed range.

We specified weakly informative priors for all model parameters. Intercept parameters for the *as* and *to be* equations were given Normal(0, 1.5) priors. Fixed-effects coefficients for 'genre' were specified with Normal(0, 1) priors. The standard deviations of the spline coefficients, which control the wiggliness of each genre-specific smooth, were given half-Student-t(3, 0, 2) priors. This regularization is especially beneficial for genres with sparse data, such as the TV/movie genre.

The model was fitted with four Markov chains, each run for 2,000 iterations (1,000 warmup), yielding 4,000 posterior draws. The target acceptance rate was set at 0.95 and the maximum tree depth to 12. The model produced no divergent transitions, and all R-hat values were below 1.01, indicating reliable convergence. Bulk and tail effective sample sizes (ESS) were adequate across all parameters (minimum bulk ESS > 1,200; minimum tail ESS > 1,500). Visual inspections of trace plots indicated good mixing, and posterior predictive checks showed that the observed data aligned well with the predicted data, confirming the validity of the model's predictions. Posterior predictive checks, R script and data are available in supplementary materials.

Table 5 reports the full posterior summaries for all model parameters. We report posterior means, the posterior standard deviation, and 95% credible intervals (lower and upper CrIs). A difference is considered statistically credible when the 95% CrI excludes zero.

**Table 5.** Summary of multinomial Bayesian GAM

| Type | Parameter | Estimate | Estimate Error | Lower 95% CI | Upper 95% CI |
|---|---|---|---|---|---|
| Population-level effects | as: Intercept | −0.909 | 0.025 | −0.960 | −0.861 |
| | to be: Intercept | −1.703 | 0.033 | −1.767 | −1.639 |
| | as: fic | −1.383 | 0.045 | −1.470 | −1.293 |
| | as: mag | −0.950 | 0.042 | −1.032 | −0.866 |
| | as: news | −1.286 | 0.120 | −1.514 | −1.043 |
| | as: tvm | −1.616 | 0.640 | −2.557 | −0.143 |
| | to be: fic | −1.126 | 0.055 | −1.234 | −1.018 |
| | to be: mag | −0.605 | 0.048 | −0.701 | −0.511 |
| | to be: news | −0.650 | 0.096 | −0.838 | −0.456 |
| | to be: tvm | −1.384 | 0.486 | −2.358 | −0.382 |
| | as: s(decade) × acad | 0.929 | 0.895 | −0.798 | 2.664 |
| | as: s(decade) × fic | −1.579 | 0.968 | −3.443 | 0.326 |
| | as: s(decade) × mag | −0.947 | 0.931 | −2.797 | 0.859 |
| | as: s(decade) × news | −0.117 | 0.988 | −2.094 | 1.876 |
| | as: s(decade) × tvm | −0.279 | 0.981 | −2.181 | 1.631 |
| | to be: s(decade) × acad | −0.256 | 0.881 | −1.986 | 1.529 |
| | to be: s(decade) × fic | 0.452 | 0.860 | −1.262 | 2.064 |
| | to be: s(decade) × mag | −0.121 | 0.900 | −1.846 | 1.637 |
| | to be: s(decade) × news | 0.605 | 0.853 | −1.305 | 2.059 |
| | to be: s(decade) × tvm | 0.173 | 1.011 | −1.750 | 2.142 |
| Smooth terms | SD as: s(decade) × acad | 3.344 | 1.050 | 1.916 | 6.034 |
| | SD as: s(decade) × fic | 1.315 | 0.678 | 0.405 | 2.997 |
| | SD as: s(decade) × mag | 2.493 | 0.899 | 1.257 | 4.733 |
| | SD as: s(decade) × news | 2.884 | 1.160 | 1.238 | 5.577 |
| | SD as: s(decade) × tvm | 1.930 | 1.491 | 0.098 | 5.739 |
| | SD to be: s(decade) × acad | 1.994 | 0.787 | 0.871 | 3.783 |
| | SD to be: s(decade) × fic | 0.984 | 0.592 | 0.268 | 2.513 |
| | SD to be: s(decade) × mag | 2.601 | 0.893 | 1.336 | 4.751 |
| | SD to be: s(decade) × news | 0.685 | 0.711 | 0.024 | 2.601 |
| | SD to be: s(decade) × tvm | 1.934 | 1.272 | 0.398 | 5.319 |

On the log-odds scale (with 'zero' as reference and academic prose as the reference genre at 1920), the intercept for *as* was −0.91 (95% CrI: [−0.96, −0.86]) and for *to be* was −1.70 (95% CrI: [−1.77, −1.64]), confirming that both variants were substantially less frequent than the zero variant in academic prose at the centering point. In fact, all genre coefficients for both *as* and *to be* were negative, indicating that every non-academic genre showed lower odds of selecting *as* or *to be* relative to zero, compared to academic prose.

Figure 4 plots the posterior mean predicted probabilities yielded from the above discussed model, and illustrates the diachronic change undergone in *consider* complementation across the five genres.

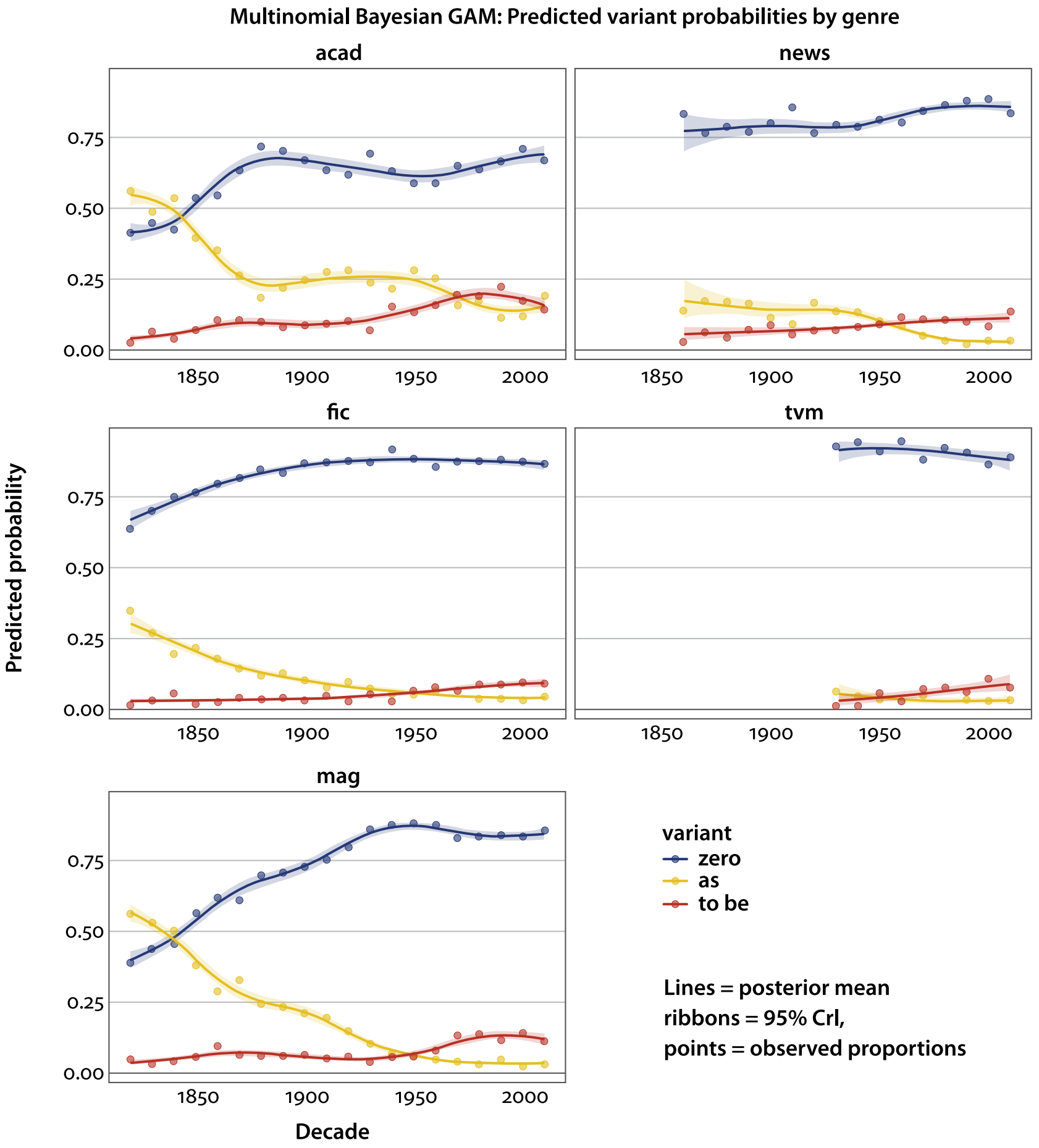

**Figure 4.** Posterior mean predicted probabilities with 95% CrIs and observed proportions overlaid as points

Notably, as Figure 4 reveals, there is a telling asymmetry across the various genres at the onset of the diachronic analysis (1820), whereby academic prose and magazine texts gravitate towards *as*-usage, and, in contrast, fiction favours the zero variant. The difference between academic prose and magazines is not statistically credible for either variant (zero: +1.6 points, 95% CrI: [−2.8, 6.1]; *as*: −2.1 points, 95% CrI: [−6.8, 2.7]), whereas the difference between academic prose and fiction is substantial and credible (zero: −25.3 points, 95% CrI: [−29.9, −20.4]; *as*: +24.2 points, 95% CrI: [19.2, 28.9]), with fiction strongly favouring the zero variant at the expense of *as*. At this point in time, the *to be* variant is still very much incipient (Nevalainen & Raumolin-Brunberg, 2016: 54–55) and only sparsely documented (acad: $N = 11$, 2.6%; mag: $N = 28$, 4.84%; fic: $N = 5$, 1.52%).

Taken together, the zero variant appears to have undergone its most rapid phase of diffusion between approximately the 1840s and the 1900s, after which it stabilized as the dominant form across all genres. Meanwhile, the *as*-complement entered a steady process of obsolescence (Rudnicka, 2021) and, synchronically, it constitutes a marginal variant across virtually all the analyzed genres, with the exception of academic prose. Most strikingly, the *to be* variant, though remaining at very low frequencies throughout the period, shows signs of incipient growth from the mid-19th century onwards. This emergent pattern suggests an ongoing shift, albeit at its early stages.

In Figure 5, we examine how the degree of genre differentiation for each variant evolves over time. For each genre × decade combination, we computed posterior expected probabilities of each variant using the posterior_epred function. Genre contrasts were computed for all pairwise combinations of genres at every available decade, yielding posterior distributions of probability differences. Pairs involving genres not yet represented in the corpus at a given decade were excluded (TV/movie before the 1930s; newspapers before the 1860s). For each contrast, we report posterior means, 95% CrI, and the posterior probability that each difference is positive. We summarize the results visually in a diachronic pairwise contrast heatmap (Figure 5), which displays the mean posterior probability difference for every genre pair across decades, faceted by variant. The full table is available in the supplementary materials (Table 6). A positive value (red) indicates that the first genre in the pair uses the variant more, and a negative value (blue) indicates that the second genre uses it more. Cells marked with an asterisk denote contrasts where the 95% CrI excludes zero.

Figure 5 reveals two main patterns. First, the zero and *as*-variants have historically been highly genre-specific, as indicated by the dense clustering of asterisks and deep colour saturation in the earlier decades. Over time, however, these differences attenuate markedly: by the late twentieth century, most pairwise contrasts for zero and *as* have faded to near-white, reflecting the convergence visible in Figure 4. The exception here is academic prose, which consistently exhibits lower probabilities of zero usage across the entire timespan, and higher probabilities of the *as* and *to be* variants. We discuss this in more detail below. Second, concerning the *to be* variant, genre differences for *to be* are modest in the early decades but become increasingly widespread from the late 19th century onwards, with credible pairwise contrasts persisting well into the 2010s, as hinted at earlier. Unlike zero and *as*, there is little sign of convergence. Reading Figures 4 and 5 together, this pattern appears to be driven by the expansion of *to be* in academic prose – where it grew from negligible levels to a stable share of around 15–18% – while remaining marginal in fiction and TV/movie dialogue throughout the observed period. Newspapers and magazines occupy an intermediate position, suggesting that the diffusion of *to be* has been limited primarily to academic writing, but with modest spread in other informational genres.

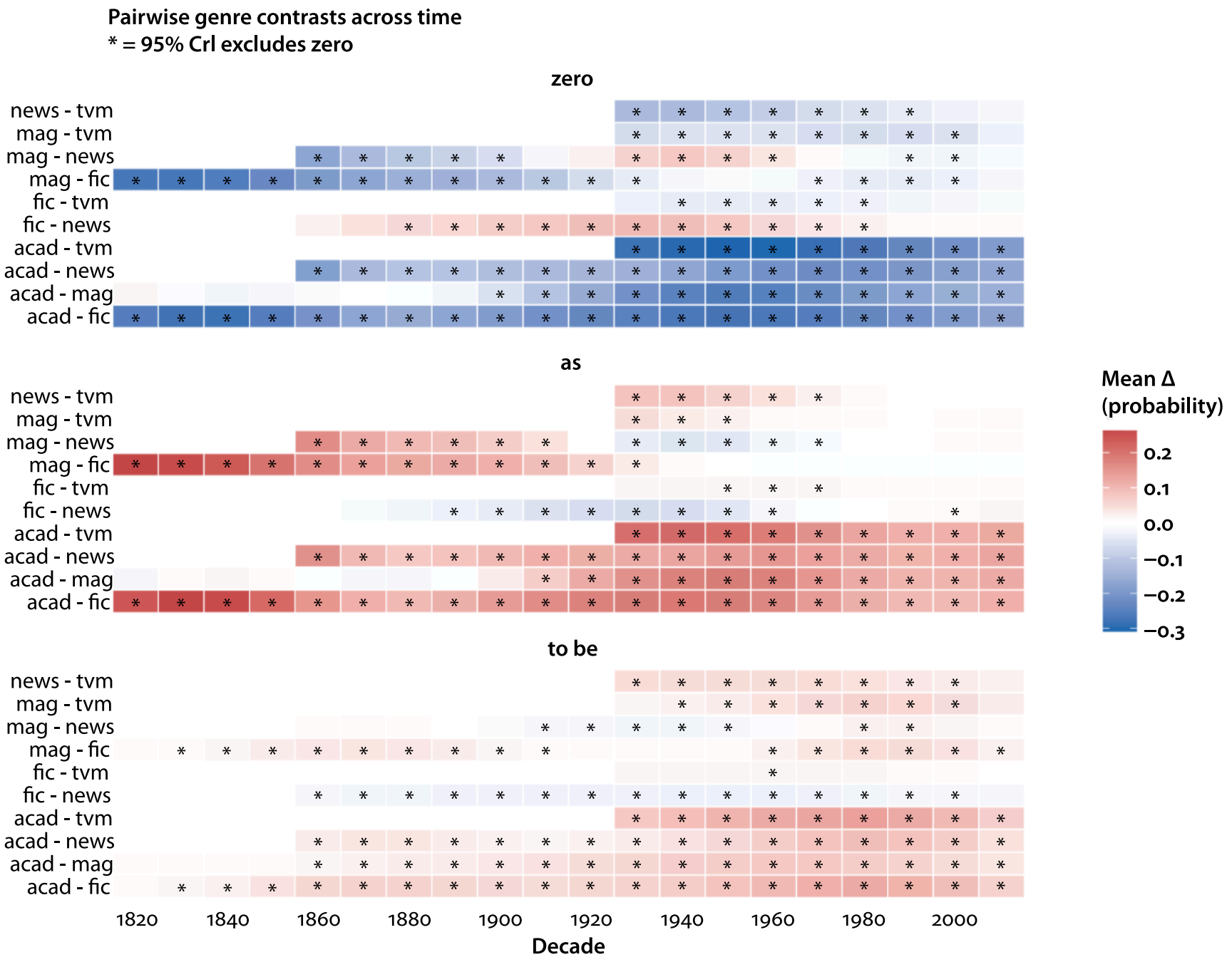

**Figure 5.** Diachronic contrast matrix of pairwise genre contrasts in predicted variant probabilities across time

Importantly, these findings could be viewed as pointing to a cyclical trajectory of change, most clearly visible in academic prose: from a probable earlier state dominated by *as*, through a period of *as*/zero variation, to zero dominance, and now toward potential zero/*to be* coexistence. While reductive processes eliminating redundant morphosyntactic material are well-documented in the literature

(Bybee, 2002, 2006, 2010; Jaeger, 2010; Levshina, 2022; Lorenz, 2013; Rickford et al., 1995), the *consider* construction thus appears to have undergone reduction (loss of *as* into zero) followed by a renewal of explicit marking through a different morphosyntactic strategy (*to be*). Such enhancement patterns remain understudied compared to their reductive counterparts, raising questions about the functional motivations driving increased explicitness in grammatical structure (for important work in this direction, see Levshina, 2022; Marttinen Larsson, 2025).

Crucially, reduction and enhancement follow systematically different paths through the genre space. The reductive shift toward zero originated in genres oriented toward narrative and dialogue — fiction and newspapers led, and TV/movie transcripts followed on entering the corpus — while academic prose resisted it most strongly, retaining the *as*-complement even as it declined elsewhere. The *to be* variant shows the inverse profile: its diffusion remained largely confined to academic prose, with only modest uptake in newspapers and magazines and persistent resistance in fiction and TV/movie dialogue.

If we accept that narrative and conversational genres such as fiction and TV/movie dialogues are characterised by a higher degree of informality than informational genres such as academic prose, news, and magazines (see Heylighen & Dewaele, 1999: 15; see also Biber, 1988 on involved versus informational registers), then this asymmetry suggests a principled relationship between register formality and the direction of morphosyntactic change. Drawing on efficiency-based accounts of grammatical variation (Hawkins, 2014; Levshina, 2022; Marttinen Larsson, 2025), we propose that these patterns reflect competing communicative pressures. Informal registers prioritise economy, favouring reduction of predictable material through *zero*-marking. Formal registers, by contrast, prioritise explicitness, favouring enhancement that aids comprehension through increased morphosyntactic specification. The incipient shift toward *to be* in academic discourse thus represents not redundancy but precision, reflecting the communicative priorities of formal written genres where clarity takes precedence over brevity.

On the basis of these findings, we tentatively advance a broader hypothesis: that enhancement and reduction are not merely different processes but are functionally anchored to different usage contexts. Reduction, as has recurrently been documented, tends to originate and often remain specialised in informal registers (Lorenz 2013; Marttinen Larsson, 2024). Enhancement, we suggest, follows the inverse path — originating in formal registers and oftentimes showing limited diffusion beyond them. That the *to be* expansion has not generalised across genres lends initial support to this view, though further evidence from other constructions and language varieties would be needed to evaluate its generality (Marttinen Larsson et al., forthcoming).

Overall, these findings confirm that the pipeline yields novel linguistic insight at a scale that was previously impractical for manual annotation. We now turn to the broader methodological implications of our study.

## 5. Discussion and conclusion

The present study set out to develop a scalable pipeline for automatic corpus annotation using large language models, addressing the methodological bottleneck that manual approaches pose for linguistic research. Our four-phase framework successfully processed the complete COHA dataset at high accuracy and speed, while revealing previously undocumented diachronic patterns in the English *consider* construction. We now turn to the broader implications of these results, examining questions of generalisability, research efficiency, accessibility, ethical considerations, and the evolving role of LLM-assisted methods in corpus linguistics.

The temporal and financial costs of our workflow merit emphasis. By contrast with the 58.6 hours and $104 reported above, manual annotation of the same dataset would require, at a conservative 10 seconds per line, roughly 400 hours of focused effort – realistically multiple weeks of full-time work over several months, with the attendant costs in salary and forgone research, or abandoning the undertaking altogether (see Section 1).

Beyond time savings, the approach restructures how intellectual labour is distributed: rather than executing repetitive, fatigue-prone classifications, the researcher invests concentrated effort upfront in prompt design and validation, after which the API applies those decisions consistently across the dataset without attention drift or inter-coder reliability concerns. The associated cost represents substantial value relative to research-assistant wages, and we argue that API fees should be normalised as legitimate budget line items in grant proposals, on a par with corpus licenses and computational resources. As the field adopts these methods, funding bodies must recognise API access as essential research infrastructure.

Beyond cost considerations, accessibility extends to the technical skills required for implementation. A crucial advantage of this pipeline lies in its openness to researchers without specialised technical training. Unlike supervised machine learning approaches requiring programming expertise, training data curation, and model development skills, our workflow operates primarily through natural language interaction. The most technically demanding component, namely API integration, requires only basic Python scripting that can be adapted from provided templates. For linguists already comfortable with corpus query languages or statistical software, this represents a modest additional skill threshold. More importantly, preliminary prompt development and validation can occur entirely through conversational interfaces before any coding becomes necessary, allowing researchers to validate the approach's viability for their specific annotation task before investing in technical implementation. This accessibility has significant implications for research equity across institutions. Sophisticated annotation has traditionally required either substantial funding for research assistants or access to computational linguists for custom tool development. Our approach requires neither. This democratisation particularly benefits junior scholars, researchers at teaching-focused institutions, and linguists working on understudied languages or constructions where existing NLP tools provide limited support.

This accessibility, however, must be balanced against important considerations about responsible data practices. The use of commercial LLM APIs for corpus annotation raises questions about data licensing and intellectual property. Our COHA license explicitly prohibited data sharing with third parties, including LLM providers. OpenAI's data-sharing-disabled configuration addressed this requirement, ensuring corpus materials would not enter training datasets. However, licensing terms vary significantly across corpora, and researchers must verify that their use complies with specific access agreements. For corpora with restrictive licenses, API-based processing may preclude certain approaches or necessitate negotiations with corpus creators for expanded usage rights. Beyond contractual obligations, broader ethical considerations emerge around data practices in LLM-assisted research. Researchers must weigh questions about where and how their data are processed, what metadata might be retained, and what assurances exist about long-term data handling. As LLM technologies evolve and corporate policies shift, maintaining alignment between research practices and ethical standards will require ongoing vigilance. We recommend that corpus linguists develop explicit guidelines for LLM-assisted annotation addressing data sovereignty, reproducibility, and transparency.

Another ethical dimension concerns the environmental impact of LLM use. Since the seminal work of Strubell et al. (2019), the NLP community has become increasingly attentive to the energy consumption and carbon emissions associated with training and deploying large models. Lifecycle analyses of major LLMs confirm that these costs are non-trivial, particularly at the training stage (Luccioni et al., 2023). However, for applications like ours that rely exclusively on inference via pre-trained models, the per-query environmental footprint is substantially smaller. Indeed, a recent comparative lifecycle assessment found that LLM inference consumes 40 to 150 times less energy, water, and carbon than equivalent human labour for the same text-processing output (Ren et al., 2024). While such ratios will vary by task and deployment context, they suggest that LLM-assisted annotation may carry a lighter environmental footprint than the hundreds of hours of manual work it replaces. Nevertheless, as the field adopts these methods at scale, aggregate impacts warrant attention, and we encourage the corpus linguistics community to advocate for transparency from API providers regarding energy consumption and to develop best practices that balance methodological innovation with ecological responsibility.

Acknowledging these practical and ethical dimensions leads naturally to recognising the method's inherent limitations. While our results demonstrate high accuracy on the *consider* annotation task, important boundary conditions constrain the pipeline's applicability. First, classification accuracy depends critically on prompt quality. Tasks requiring extremely nuanced semantic distinctions, extensive discourse context, or specialised domain knowledge may prove challenging even with careful prompt engineering. Second, the approach works best for annotation schemes with relatively clear decision criteria. Highly subjective or impressionistic coding systems may resist systematisation. Third, although we achieved 98%+ accuracy, even small error rates compound across very large datasets, potentially introducing subtle biases into downstream analyses. Researchers must assess whether their analytical methods are robust to this level of noise. Most fundamentally, LLM-assisted annotation cannot replace deep linguistic expertise. The pipeline requires human researchers to conceptualise the annotation task, recognise relevant edge cases, interpret validation results, and take responsibility for data quality. LLMs serve as powerful tools for executing classification decisions at scale, but they cannot substitute for the theoretical knowledge and analytical judgment that corpus linguists bring to their research.

A related consideration concerns the optimal level of prompt detail. Fuoli et al. (2026) compared three approaches to LLM-assisted metaphor identification: retrieval-augmented generation (RAG), where models were provided with a detailed external codebook; prompt engineering strategies including zero-shot, few-shot, and Chain of Thought; and fine-tuning. RAG performed worst overall, while Chain of Thought prompting and fine-tuning achieved the best results – approaches that also involve substantial guidance, albeit integrated directly into the prompt or model rather than retrieved from external documents. Our results align with this pattern: detailed instructions improve performance when embedded within a well-structured prompt using XML tags and explicit reasoning steps, rather than offloaded to supplementary materials.

Looking forward from these limitations, our case study focused on English syntactic variation, but we believe that the methodology extends readily to other linguistic phenomena, languages, and corpora. Semantic annotation, pragmatic tagging, discourse structure analysis, and sociolinguistic variation all present viable applications. Multilingual annotation represents a particularly promising direction, as contemporary LLMs demonstrate strong capabilities across numerous languages. The approach may prove especially valuable for low-resource languages where annotated training data for supervised methods remain scarce. As LLM capabilities continue advancing, we anticipate further efficiency gains and accuracy improvements. However, the field must balance enthusiasm for technological solutions with critical evaluation of their limitations. The most productive path forward likely involves hybrid approaches where LLM-assisted annotation handles scalable components while human expertise addresses cases requiring deeper interpretation. The goal is not to eliminate human involvement but to deploy it strategically where it adds greatest value.

To conclude, our study contributes to the growing literature showing that contemporary LLMs can perform linguistically sophisticated annotation at scales previously unattainable, opening research questions that manual methods rendered impractical. By combining prompt engineering with systematic validation, corpus linguists can harness these tools to extend empirical inquiry while maintaining the methodological rigour that defines the field. As we navigate this technological moment, the central question is not whether to adopt LLM-assisted methods, but how to integrate them thoughtfully into corpus research practice.

## Address for correspondence

Matti Marttinen Larsson
Department of Romance Studies and Classics
Stockholm University
Sweden

matti.marttinen@su.se
https://orcid.org/0000-0002-6224-7872

## Co-author information

Cameron Morin
Approches Linguistiques Théoriques,
Appliquées et Expérimentales: Langues et
cultures connectées
Paris Cité University

cameron.morin@u-paris.fr

## Publication history